\documentclass[10pt,twocolumn,letterpaper]{article}

\usepackage[pagenumbers]{wacv} 

\usepackage{xcolor}
\usepackage[normalem]{ulem}

\usepackage[utf8]{inputenc}

\usepackage{graphicx}
\usepackage{multirow} 

\definecolor{wacvblue}{rgb}{0.21,0.49,0.74}
\usepackage[pagebackref,breaklinks,colorlinks,allcolors=wacvblue]{hyperref}
\usepackage{stmaryrd}
\usepackage{amsmath}
\usepackage{graphicx}
\usepackage{dsfont}

\def\wacvPaperID{955} 
\def\confName{WACV}
\def\confYear{2026}

\newcommand\blfootnote[1]{
    \begingroup
    \renewcommand\thefootnote{}\footnote{#1}
    \addtocounter{footnote}{-1}
    \endgroup
}

\title{VISTA: A Vision and Intent-Aware Social Attention Framework for Multi-Agent Trajectory Prediction}

\author{Stephane Da Silva Martins, Emanuel Aldea and Sylvie Le Hégarat-Mascle \\
SATIE - CNRS UMR 8029 \\
Paris-Saclay University, France \\
\tt\small\{stephane.da-silva-martins, emanuel.aldea, sylvie.le-hegarat\}@universite-paris-saclay.fr
}

\begin{document}
\maketitle
\begin{abstract}

Multi-agent trajectory prediction is a key task in computer vision for autonomous systems, particularly in dense and interactive environments. Existing methods often struggle to jointly model goal-driven behavior and complex social dynamics, which leads to unrealistic predictions. In this paper, we introduce \textbf{VISTA}, a recursive goal-conditioned transformer architecture that features (1) a cross-attention fusion mechanism to integrate long-term goals with past trajectories, (2) a social-token attention module enabling fine-grained interaction modeling across agents, and (3) pairwise attention maps to show social influence patterns during inference. Our model enhances the single-agent goal-conditioned approach into a cohesive multi-agent forecasting framework. In addition to the standard evaluation metrics, we also consider trajectory collision rates, which capture the realism of the joint predictions. Evaluated on the high-density MADRAS benchmark and on SDD, VISTA achieves state-of-the-art accuracy with improved interaction modeling. On MADRAS, our approach reduces the average collision rate of strong baselines from $2.14\%$ to $0.03\%$, and on SDD, it achieves a $0\%$ collision rate while outperforming SOTA models in terms of ADE/FDE and minFDE. These results highlight the model’s ability to generate socially compliant, goal-aware, and interpretable trajectory predictions, making it well-suited for deployment in safety-critical autonomous systems.

\end{abstract}
    
\section{Introduction}
\label{sec:intro}

Multi-agent trajectory prediction has become a critical task for various autonomous systems, including self-driving cars, social robots, and intelligent surveillance\blfootnote{This study was supported by the OLICOW project, funded by the French National Research Agency (ANR) grant ANR-22-CE22-0005.} \kern-0.7em systems~\cite{survey}. The main goal of this task is to anticipate the future positions and movements of agents, such as pedestrians, vehicles, and cyclists, based on their historical trajectories and contextual surroundings. The surrounding environment comprises both static components (e.g., buildings, roads, and fixed obstacles) and dynamic entities (e.g., nearby moving pedestrians and vehicles).

Early approaches to pedestrian trajectory prediction were inspired by physics-based models, notably drawing on Newton’s third law to conceptualize pedestrian dynamics through social forces~\cite{social_helbing}. These approaches modeled pedestrian interactions using attractive and repulsive forces, accounting for the impact of static environmental features~\cite{lohner}. However, such physics-based models often struggled to handle complex, nuanced human interactions and behaviors, as these interactions often entail subtle social cues and contextual factors not easily captured by explicit force formulations.

To overcome these limitations, recent advances have focused on data-driven approaches leveraging deep learning techniques, which have significantly improved predictive performance. Data-driven methods are particularly effective at capturing the intricate spatial and temporal relationships among pedestrians. Notable approaches include graph neural networks (GNNs)~\cite{bae2023set, bae2022learning}, recurrent neural networks (RNNs)~\cite{syed2023semantic, alahi2016social}, conditional variational autoencoders (CVAEs)~\cite{Mangalam2020PECNet, yuan2021agentformer}, and models using transformer-based attention mechanisms~\cite{cheng2023gatraj, tran2021goal}.

However, current approaches often struggle to simultaneously capture both the goal-oriented behaviors and the nuanced social dynamics of pedestrians. Existing models can either adequately model agent interactions but lack long-term goal integration or focus heavily on individual goals without adequately accounting for dynamic interactions, resulting in socially unrealistic predictions. Furthermore, interpretability remains limited in most State-Of-The-Art~(SOTA) models, complicating their application in safety-critical scenarios where understanding the rationale behind decisions is crucial.


In this paper, we propose \textbf{VISTA}, a recursive goal-conditioned multi-agent model which decouples the prediction of 
destination goals from local trajectory generation, while taking into account the scene context and agent dynamics. This structured separation makes it possible to model agents' intention, generate flexible trajectories and improve the interpretability of the interactions between agents.

Crucially, VISTA builds upon the single-agent, goal-conditioned paradigm introduced in~\cite{YNet2021}, and extends it into a joint, multi-agent framework. By integrating
goal conditioning with recursive social attention, our model jointly refines trajectories while capturing mutual influences (e.g., collision avoidance) in dense scenes. While each of these components has been explored individually, VISTA is the first to unify them within an end-to-end framework that achieves both high accuracy and very low collision rates.

VISTA addresses the aforementioned challenges through three key design innovations:

\begin{itemize}    
    \item \textbf{Goal–trajectory fusion.} A dedicated cross‑attention block combines the goal heatmap embedding with the past‑trajectory embedding at each recursive step, ensuring that the long‑term goal is effectively integrated into the past trajectory before predicting the next time step position.
    \item \textbf{Social‑token attention.} Each agent
    is represented as a learnable token within a multi‑head social‑attention layer, enabling information sharing between agents. This allows the model to jointly reason about mutual influence, coordination, and collision avoidance at every time step.
    \item \textbf{Pairwise attention maps.} The social‑attention mechanism produces stepwise attention matrices 
    that explicitly indicate which pedestrian attends to which neighbor. These maps reveal the pairwise influence structure steering each prediction and provide a clear window into the model’s social reasoning.
\end{itemize}

Evaluated on SDD and the high‑density MADRAS benchmark, VISTA achieves the SOTA accuracy with below $0.1\%$ collision rate, demonstrating both performance and social compliance. On the high‑density MADRAS benchmark (nine crowd scenes), VISTA reduces the average collision rate from $2.14\%$ for MART to $0.03\%$, while also ranking among the top models in terms of accuracy. On SDD, our model improves ADE/FDE by $24\% / 27\%$ and achieves a perfect $0\%$ collision rate, outperforming ten recent SOTA baselines in minFDE.

\textbf{Our main contributions are:}
\begin{itemize}

    \item A novel end-to-end approach, in a unified framework, that integrates goal heatmap conditioning, recursive multi-agent refinement, and interpretable pairwise attention:
    using both goal–trajectory fusion and social attention over learnable agent tokens, our model extends the single-agent goal-conditioned paradigm into a coherent multi-agent forecasting formulation.
    \item Strong and consistent performance across diverse benchmarks, supporting the generality of the proposed architecture and its applicability to both sparse and high-density crowd scenarios.
    \item 
    Qualitative and quantitative evidence that pairwise attention maps capture agent-to-agent influence patterns: collision rate serves as a key evaluation metric, highlighting our model’s potential in safety-critical settings.
\end{itemize}

\section{Related Work}
\label{sec:formatting}

\paragraph{Multi-agent trajectory prediction.}
Early research combined hand-crafted physical or social rules with probabilistic filters to forecast movement in crowds~\cite{social_helbing,Berndt2008HMMVehicle}. However, deep sequence models replaced these heuristic-based approaches. Social LSTM effectively captured crowd-level temporal dependencies~\cite{alahi2016social}, while Social Attention enhanced trajectory representation by incorporating interaction-aware attention mechanisms~\cite{Vemula2018SocialAttention}. Generative approaches then emerged in the field. Social GAN introduced adversarial learning to account for the inherently multimodal nature of future trajectories~\cite{Gupta2018SocialGAN}. PECNet subsequently improved long-horizon forecasting with a two-stage CVAE design along with social pooling, yielding more sample-efficient and realistic long-range predictions~\cite{Mangalam2020PECNet}. More recently, transformer-based forecasting models, such as AgentFormer, STGT, and Social-SSL, have begun to jointly reason over time, space, and higher-order interactions, setting new SOTA benchmarks~\cite{yuan2021agentformer,syed2021stgt,tsao2022social}. The latest diffusion-based methods have narrowed the realism gap even further by sampling trajectories from powerful generative priors~\cite{Li2023TrajectoryDiffusion}.

\paragraph{Goal-conditioned prediction.}
Conditioning on agents’ intended destinations, i.e. `goals', tightens the learning signal and yields more coherent joint futures. PECNet pioneered goal conditioning for pedestrians~\cite{Mangalam2020PECNet}. Y-Net, which relies on a U-Net-style encoder–decoder, first produces a heatmap of plausible destinations per agent and samples a goal. This goal is then used to generate successive heatmaps of future positions, yielding a refined, scene-coherent trajectory. SwYn-Net~\cite{SwYnNet2024} leverages Swin-Transformer shifted-window attention to combine local and global context within the Y-Net architecture. Goal-SAR~\cite{GoalSAR2022} introduces a lightweight self-attentive recurrent backbone paired with a scene-aware goal module, slightly surpassing Y-Net while using a simpler architecture. Y-Net and Goal-SAR are primarily single-agent predictors; in contrast, VISTA takes the social context into account. Recently, Di-Long~\cite{das2024distilling} further extends goal conditioning to multi-agent long-term forecasting through a short-to-long knowledge distillation framework but requires an externally provided social mask to delimit interactions. 
Di-Long and PECNet require a social mask to enable the model to account for interactions. In contrast, VISTA infers interactions autonomously and employs attention maps to interpret these relationships.

\paragraph{Attention-based modeling of social context.}

Alongside goal-driven methods, a vast literature has explored how to capture social interactions through attention mechanisms. Early approaches introduced graph-based attention. Social-BiGAT~\cite{kosaraju2019socialbigat} modeled pedestrians as nodes in a graph and applied graph attention networks to capture pairwise influences. Social-STGCNN~\cite{mohamed2020socialstgcnn} combined temporal convolutions with spatial attention over interaction graphs. STAR~\cite{yu2020star} employed relational attention mechanisms to dynamically focus on the most relevant neighbors in crowded scenarios. 
Beyond graph-based models, Trajectron++~\cite{salzmann2020trajectronpp} integrated an attention mechanism within a conditional variational framework to generate multimodal and diverse trajectory samples. 
In recent years, transformer-based models have become dominant in trajectory forecasting. AgentFormer~\cite{yuan2021agentformer} introduced a spatio-temporal transformer to jointly encode past trajectory and agent interactions. SceneTransformer~\cite{ngiam2021scenetransformer} extended this approach by integrating global scene encoding with local agent-to-agent attention. GroupNet~\cite{liang2022groupnet} explicitly modeled group-level behaviors through hierarchical attention. MemoNet~\cite{jia2022memonet} incorporated memory-based attention to capture long-term dependencies. MID~\cite{park2022mid} proposed interpretable attention maps that highlight potential causal interactions between agents. 
More recently, Social-Transformer~\cite{zhong2023socialtransformer} and D2-TPred~\cite{chen2023d2tpred} leveraged multi-scale and disentangled attention to improve robustness in dense crowds.
Despite recent advances, modeling social interactions remains challenging. Most attention-based methods face scalability limitations, exhibiting quadratic complexity as the number of agents increases. Furthermore, the coupling between agent intent and interaction remains insufficient. To address these issues, we propose VISTA, a scalable, goal-conditioned model, presented in the next section.

\section{Method}
\label{sec:method}

\begin{figure*}[ht]
    \centering
    \includegraphics[width=1.\linewidth]{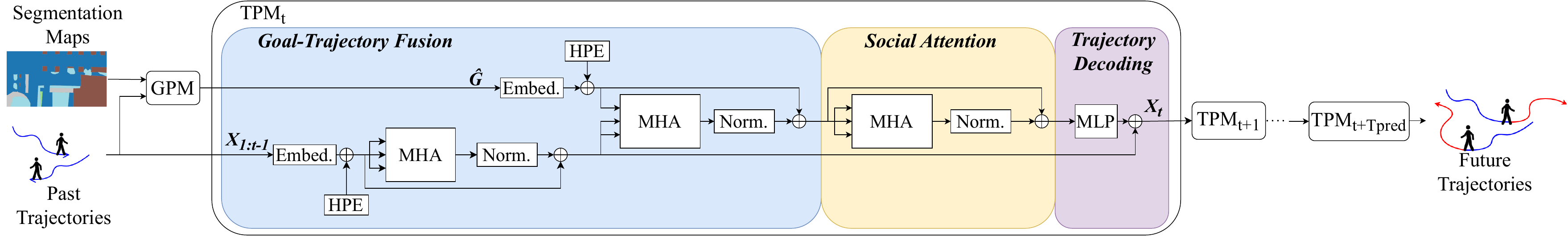}
    \caption{\textbf{Overview of the VISTA architecture.} The GPM utilizes past trajectories and a scene segmentation map to generate per-agent goal heatmaps and continuous goals. Subsequently, the TPM applies hybrid positional encoding, integrates the goal token with past trajectory embeddings using cross-attention, and performs social self-attention across agents. Finally, the TPM recursively decodes displacements to generate future trajectories. MHA denotes Multi-Head Attention, Embed. denotes Embedding, Norm. denotes Normalization, and MLP denotes Multi-Layer Perceptron. 
}
    \label{fig:enter-label}
\end{figure*}
\subsection{Problem Definition and Overview
}

Let $\mathcal{S}$ denote a scene of interest represented by a static image $\mathcal{I} \in \mathbb{R}^{H \times W \times C}$, where $H$, $W$, and $C$ are the image height, width, and number of channels, respectively. In this scene, $N$ agents are observed over $T_{\text{obs}}$ time steps
, with their spatial positions given by:
\begin{center}
    $\mathbf{X}_{1:T_{obs}} = \left\{ \mathbf{x}_t^i \in \mathbb{R}^2 ~\middle|~ t \in \llbracket 1, T_{\text{obs}} \rrbracket,~ i \in \llbracket 1, N \rrbracket \right\}.$
\end{center}

The objective of a multi-agent trajectory model is to predict future trajectories over $T_{\text{pred}}$ time steps:
\begin{center}
    $\hat{\mathbf{Y}} = \left\{ \hat{\mathbf{y}}_t^i \in \mathbb{R}^2 ~\middle|~ t \in \llbracket T_{\text{obs}}+1, T_{\text{pred}} \rrbracket,~ i \in \llbracket 1, N \rrbracket \right\}.$
\end{center}

We develop \textbf{VISTA}, a modular and recursive trajectory prediction architecture specifically designed to jointly model \emph{long-term goal intention}, \emph{social interactions}, and \emph{spatial semantics}. 
The key idea is the extension and integration of a single-agent goal-conditioned paradigm into a joint multi-agent prediction framework, based on:

\begin{enumerate}
    \item \textbf{Goal Prediction Module} that generates dense spatial goal heatmaps from past trajectory and scene context,
    \item \textbf{Goal-trajectory fusion} that combines goal and past trajectory embeddings at each recursive step using a cross-attention mechanism, and
    \item \textbf{Social attention} that captures agent interactions using token-level attention.
\end{enumerate}

Formally, our model conditions each prediction on the past trajectory $\mathbf{X}_{1:t-1}$, the static scene image $\mathcal{I}$, and the predicted goal location $\hat{\mathbf{g}}^i$ for each agent, with the goal being defined as the final position of the trajectory. So, 
\begin{equation}
    \hat{\mathbf{y}}_t^i = \mathcal{M}(\mathcal{I},\mathbf{X}_{1:t-1},\hat{\mathbf{g}}^i ) 
\end{equation}
Such prediction is achieved by three modules described in Fig.~\ref{fig:enter-label} that recursively refine trajectory predictions over multiple steps.

Figure~\ref{fig:enter-label} highlights the sequential link between the Goal Prediction Module (GPM) and the Trajectory Prediction Module (TPM) as well as the recursive nature of TPM. Specifically, the GPM~(cf. Section~\ref{sec:GPM}) utilizes past trajectories and a scene segmentation map to generate per-agent goal heatmaps and continuous goals. Subsequently, the TPM~(cf. Section~\ref{sec:TPM}) applies hybrid positional encoding, integrates the goal token with past trajectory embeddings using cross-attention, and performs social self-attention across agents.


\subsection{Goal Prediction Module}
\label{sec:GPM}
The GPM captures agent intent through long-term destination prediction. Unlike standard trajectory-only conditioning, we explicitly leverage scene semantics in addition to previous trajectory. The spatial context is represented through a $D$-class segmentation map computed using a pretrained backbone $f_\text{seg}$:
\begin{equation}
    \mathcal{F}_\text{seg} = f_\text{seg}(\mathcal{I}) \in \mathbb{R}^{H \times W \times D}.
\end{equation}

In our model, the per-agent heatmaps are generated by a U-Net–based decoder, conditioned on the past trajectory and the segmentation $\mathcal{F}_\text{seg}$:
\begin{equation}
    \mathcal{H}_\text{goal}^i = f_\text{goal}(\mathbf{x}_{1:T_{\text{obs}}}^i, \mathcal{F}_\text{seg}).
\end{equation}

Note that to bridge the gap between discrete heatmaps and continuous goal coordinates, we apply a differentiable \textit{softargmax} operator $\sigma$:

\begin{equation}
    \hat{\mathbf{g}}^i = \sigma(\mathcal{H}_\text{goal}^i) \in \mathbb{R}^2.
\end{equation}

This explicit goal representation serves as a global guide for recursive decoding. To provide multiple plausible goals, we adopt the Test-Time Sampling Trick (TTST)~\cite{YNet2021} to our setting, applying large-scale sampling followed by K-means clustering to obtain a diverse yet compact set of candidate goals.  

\subsection{Trajectory Prediction Module}
\label{sec:TPM}
The Trajectory Prediction Module (TPM) forms the core of VISTA. It takes as input the predicted goals and recursively generates multi-agent trajectories. Unlike prior models that either treat goals as static priors (Di-Long) or as decoupled coordinates (PECNet), our TPM fuses goals with trajectory history through cross-attention, and refines these representations via social-token attention across agents. This recursive decoding loop keeps each prediction step grounded in both long-term intent and ongoing social interactions, yielding socially compliant and scene-aware trajectories.

\subsubsection{Hybrid Positional Encoding}
We adopt a Hybrid Positional Encoding (HPE) strategy that combines fixed sinusoidal encodings with learnable offsets, following common practice in sequence modeling. This balances inductive temporal structure with adaptive flexibility. At time step $t$, the HPE is defined as:
\begin{equation}
\mathrm{HPE}(\mathbf{e}_t) = \mathbf{e}_t + \mathbf{p}_t^{\text{sin}} + \mathbf{p}_t^{\text{learn}},
\end{equation}
where $\mathbf{p}_t^{\text{sin}}$ is the standard sinusoidal encoding~\cite{transformer} and $\mathbf{p}_t^{\text{learn}}$ a learnable offset~\cite{bert}. This encoding is applied to both position tokens and the goal token, which is assigned a reserved position (e.g., $t = T_{\text{pred}}$).
This explicit alignment embeds the goal as part of the temporal sequence, treating it as a first-class token in the transformer rather than an auxiliary feature.

\subsubsection{Goal-Trajectory Fusion}
\label{sec:fusion}

Given the predicted goal $\hat{\mathbf{g}}^i \in\mathbb{R}^2$ for agent $i$, TPM enters a recursive decoding phase. Each past position $\mathbf{x}_t^i$ is projected into the embedding space via a linear map $\phi: \mathbb{R}^2 \rightarrow \mathbb{R}^d$, producing $\mathbf{e}_t^i$, and then added with HPE to yield a temporal position token sequence $\mathbf{E}_{1:t-1}^i$.

Then, our model explicitly conditions this sequence on the goal through cross-attention: a dedicated goal token $\mathbf{e}_g^i$ interacts with the past position tokens, ensuring that trajectory generation is continuously aware of the agent’s final intent. Formally, temporal dependencies are first modeled by self-attention:
\begin{equation}
\mathbf{T}_{1:t-1}^i = \mathrm{MHA}(\mathbf{E}_{1:t-1}^i, \mathbf{E}_{1:t-1}^i, \mathbf{E}_{1:t-1}^i),
\end{equation}
then refined via cross-attention with the goal embedding:
\begin{equation}
\mathbf{Z}_{1:t-1}^i = \mathrm{MHA}(\mathbf{T}_{1:t-1}^i, \mathbf{e}_g^i, \mathbf{e}_g^i).
\end{equation}

Finally, the feature vector from the last time step is selected as the fused trajectory representation at step $t$:

\begin{equation}
\mathbf{h}_{t-1}^i = \text{Norm}\!\left(\mathbf{Z}_{t-1}^i\right) + \mathbf{T}_{t-1}^i,
\end{equation}
where $\text{Norm}$ denotes a layer normalization. The resulting representation $\mathbf{h}_{t-1}^i \in \mathbb{R}^d$ integrates position history and long-term intent, and feeds directly into the social modeling stage.

\subsubsection{Social Attention}
To model social interactions, we apply a transformer-based self-attention module across all agents at each time step. While this enables global interaction modeling, its complexity scales as $\mathcal{O}(N^2)$ with the number of agents $N$. This is tractable for typical scenes with tens of agents, but can become costly in denser crowds. Prior works address this issue through pruning~\cite{smart}, local attention over k-nearest neighbors~\cite{hivt, mrt}, or graph-based modeling~\cite{mart}, albeit at the risk of discarding long-range dependencies. We keep full self-attention across agents for simplicity and accuracy. Its $\mathcal{O}(N^2)$ cost is tractable on our benchmarks but may limit scalability in very dense scenes. We leave efficiency improvements to future work.

Once each trajectory is aligned with its predicted goal, the next step is to refine the representations through social interactions. At prediction step $t$, the goal-aware features of all $N$ agents are gathered into a joint matrix:
\begin{equation}
\mathbf{H}_{t-1} = [\mathbf{h}_{t-1}^1, \ldots, \mathbf{h}_{t-1}^N]^\top \in \mathbb{R}^{N \times d}.
\end{equation}

Each row of $\mathbf{H}_{t-1}$ encodes the feature representation of a single agent. By stacking these rows, the model jointly considers all agents, upon which we can apply multi-head self-attention:
\begin{equation}
\mathbf{S}_{t-1} = \mathrm{MHA}(\mathbf{H}_{t-1}, \mathbf{H}_{t-1}, \mathbf{H}_{t-1}),
\end{equation}
The updated feature for agent $i$ is then directly given by the $i$-th row:
\begin{equation}
\tilde{\mathbf{h}}_{t-1}^i = \mathbf{S}_{t-1}[i] \in \mathbb{R}^d.
\end{equation}

This design introduces learnable social tokens that enable pairwise attention maps, yielding directly interpretable interaction patterns. Unlike graph-based or local attention approaches, our mechanism preserves global context without discarding long-range dependencies, which is crucial for dense crowds. 

\subsubsection{Trajectory Decoding}
At each prediction step $t \in \llbracket T_{\text{obs}}+1, T_{\text{pred}} \rrbracket$, the interaction-aware feature vector $\tilde{\mathbf{S}}_{t-1}[i]$ is fed into a multilayer perceptron (MLP) to produce a displacement vector:
\begin{equation}
\Delta \hat{\mathbf{y}}_t^i = \mathrm{MLP}(\mathbf{h}_{t-1}^i) \in \mathbb{R}^2.
\end{equation}

This displacement is added to the last predicted position of the agent, and the resulting positions are appended to their respective trajectory sequences.
\begin{equation}
\hat{\mathbf{y}}_t^i = \hat{\mathbf{y}}_{t-1}^i + \Delta \hat{\mathbf{y}}_t^i,
\end{equation}
\begin{equation}
\hat{\mathcal{Y}}_{1:t}^i = [\mathbf{x}_1^i, \ldots, \mathbf{x}_{T_{\text{obs}}}^i, \hat{\mathbf{y}}_{T_{\text{obs}}+1}^i, \ldots, \hat{\mathbf{y}}_t^i].
\end{equation}

These sequences are subsequently re-embedded and re-encoded in the next iteration. This process is repeated recursively for all time steps $t = T_{\text{obs}} + 1$ to $T_{\text{pred}}$. At each step, all agents are processed in parallel, with the transformer modules reused over time through shared weights.

By unrolling this iterative mechanism, the model generates complete predicted trajectories:
\begin{equation}
\hat{\mathcal{Y}}^i = [\hat{\mathbf{y}}_{T_{\text{obs}}+1}^i, \ldots, \hat{\mathbf{y}}_{T_{\text{pred}}}^i] \in \mathbb{R}^{(T_{\text{pred}} - T_{\text{obs}}) \times 2}.
\end{equation}

\subsection{Loss Function}

The model is optimized using a joint loss that supervises both the goal prediction and trajectory generation.

\paragraph{Goal-prediction loss.}
The Goal Prediction Module outputs a spatial probability distribution $\hat{Q} \in [0,1]^{H \times W}$ over a discretized 2D grid. As in~\cite{YNet2021}, it is trained using a binary cross-entropy (BCE) loss against a ground-truth heatmap $Q$, modeled as a normalized 2D Gaussian centered on the agent’s final ground-truth position $\mathbf{y}_{T_{\text{pred}}}^i$. The BCE loss for agent $i$ is then:
\begin{equation}
\mathcal{L}_{\text{goal}}^i = \mathrm{BCE}(\hat{Q}^i, Q^i).
\end{equation}

\paragraph{Trajectory-prediction loss.}
The Trajectory Prediction Module outputs a sequence of predicted positions $\hat{\mathbf{Y}}^i = [\hat{\mathbf{y}}_{T_{\text{obs}}+1}^i, \ldots, \hat{\mathbf{y}}_{T_{\text{pred}}}^i]$. It is trained using a mean squared error (MSE) loss against the ground-truth trajectory for future time steps $\mathbf{Y}^i = [\mathbf{y}_{T_{\text{obs}}+1}^i, \ldots, \mathbf{y}_{T_{\text{pred}}}^i]$:
\begin{equation}
\mathcal{L}_{\text{traj}}^i = \frac{1}{T_{\text{pred}} - T_{\text{obs}}} \sum_{t = T_{\text{obs}}+1}^{T_{\text{pred}}} \left\| \hat{\mathbf{y}}_t^i - \mathbf{y}_t^i \right\|_2^2.
\end{equation}

\paragraph{Joint loss.}
The overall loss is a weighted sum of both components:
\begin{equation}
\mathcal{L}_{\text{total}} = \lambda_{\text{goal}} \sum_{i=1}^N \mathcal{L}_{\text{goal}}^i + \lambda_{\text{traj}} \sum_{i=1}^N \mathcal{L}_{\text{traj}}^i,
\end{equation}
where $\lambda_{\text{goal}}$ and $\lambda_{\text{traj}}$ are hyperparameters balancing goal and trajectory.

Training both modules jointly with shared gradients enables coordinated learning of high-level intention (goal) and fine-grained motion (trajectory).
\section{Experiments}
\label{sec:experiment}

\subsection{Datasets}
For the evaluation of VISTA, we use two datasets for multi-agent trajectory prediction: the Stanford Drone Dataset (SDD)~\cite{sdd} and MADRAS~\cite{madras}.

MADRAS is a large-scale pedestrian trajectory dataset collected during the 2022 Festival of Lights in Lyon, France, using zenithal (bird's eye view) cameras. It contains over $7,000$ pedestrian trajectories recorded in high-density crowd conditions across nine different urban scenes. The dataset captures real-world human trajectories in unstructured settings, with densities reaching up to $4$ pedestrians per square meter. For MADRAS, we follow prior works that use a leave-one-out method for model training; specifically, we train the proposed model on eight subsets and test it on the rest of the subsets. This process is repeated across all possible train/test splits, and performance metrics are averaged over the nine folds to ensure robustness and comparability. 
Compared to SDD, which includes $5,232$ pedestrian trajectories captured across eight campus-like scenarios with diverse agents, MADRAS focuses exclusively on dense pedestrian flows in crowded urban environments, making it especially relevant for evaluating multi-agent trajectory predictions in high-density conditions.

For these datasets, we consider past trajectory length $T_{obs} = 8$ ($3.2$ seconds) and future trajectory length $T_{pred} = 12$ ($4.8$ seconds).

\subsection{Implementation Details} 

Our model was trained using the Adam optimizer with a learning rate of $10^{-3}$ for the SDD and MADRAS datasets, for up to $500$ epochs on GPU NVIDIA TITAN RTX with $24$ GB RAM. Our convergence scheme was that if the validation ADE did not decrease after $30$ epochs, we reduced the learning rate by a factor of $2$. Additionally, we applied early stopping to terminate training after $75$ epochs without improvement on the validation minADE. Similarly to~\cite{YNet2021}, we scaled the goal loss by $\lambda_{\text{goal}} = 10^3$ and the trajectory loss by $\lambda_{\text{traj}} = 1$. We downsampled the scene images by $4$ for SDD. The number of training data was increased by $32$ using spatial flipping and rotations in $90$° steps. We use an embedding dimension of $32$ for spatial coordinates and $8$ heads in MHA. 

In the multimodal setting, the model generates $k$ trajectories (per agent), called trajectory samples. In this work, we set $k = 20$, as in~\cite{YNet2021, mart, tutr}. In the following, the set of trajectory samples of a given agent $i$ is denoted by $\{\hat{\mathbf{Y}}^{i,j} \}_{j=1}^k$, where $\hat{\mathbf{Y}}^{i,j} = [\hat{\mathbf{y}}_{T_{\text{obs}}+1}^{i,j}, \ldots, \hat{\mathbf{y}}_{T_{\text{pred}}}^{i,j}]$ denotes the $j^{th}$ trajectory sample.

\subsection{Evaluation Metrics}
\label{sec:metrics}
We consider a set of standard metrics commonly used in trajectory forecasting: Average Displacement Error (ADE), Final Displacement Error (FDE), their multimodal variants (minADE$_k$, minFDE$_k$), Area Under the Curve (AUC), and the Collision Rate.
\\ADE measures the average $\ell_2$ distance 
between predicted and ground-truth positions over the entire prediction horizon, the $k$ generated samples and the agents:
\begin{equation}
\mathrm{ADE} = \frac{1}{N k \Delta T} \sum_{i=1}^{N} \sum_{j=1}^{k} \sum_{t = T_{\text{obs}}+1}^{T_{\text{pred}}} \left\| \hat{\mathbf{y}}_t^{i,j} - \mathbf{y}_t^i \right\|_2.
\end{equation}
\\FDE computes the $\ell_2$ distance at the final prediction step, also averaged over the $k$ generated samples and the agents:
\begin{equation}
\mathrm{FDE} = \frac{1}{N k} \sum_{i=1}^{N} \sum_{j=1}^{k} \left\| \hat{\mathbf{y}}_{T_{\text{pred}}}^{i,j} - \mathbf{y}_{T_{\text{pred}}}^i \right\|_2.
\end{equation}
The metrics minADE$_k$ and minFDE$_k$ select, among $k$ predicted trajectories, the one closest to the ground-truth in terms of ADE and FDE, respectively:
\begin{align}
\mathrm{minADE}_k &= \frac{1}{N} \sum_{i=1}^N \min_{j = 1,\ldots,k} \left( \frac{1}{\Delta T} \sum_{t = T_{\text{obs}}+1}^{T_{\text{pred}}} \left\| \hat{\mathbf{y}}_t^{i,j} - \mathbf{y}_t^i \right\|_2 \right), \nonumber\\
\mathrm{minFDE}_k &= \frac{1}{N} \sum_{i=1}^N \min_{j = 1,\ldots,k} \left\| \hat{\mathbf{y}}_{T_{\text{pred}}}^{i,j} - \mathbf{y}_{T_{\text{pred}}}^i \right\|_2.
\end{align}

Despite their popularity and as pointed out by some authors, minADE and minFDE metrics are biased toward models that favor broad sampling over the trajectory space. To mitigate this bias and provide a fairer evaluation, we adopt the AUC metric introduced in~\cite{auc}, which integrates over different numbers of trajectory samples. Specifically, for a given number of samples $K\in\llbracket 1, k\rrbracket$ usually selected as the top-$K$ trajectories, we define the expected displacement error $E_K$ as:
\begin{equation}
E_K^i = \frac{1}{T_{\text{pred}}} \sum_{j=1}^{k - K + 1}\frac{\binom{k - j}{K - 1}}{\binom{k}{K}} \left\| \hat{\mathbf{Y}}^{i,j} - \mathbf{Y}^{i} \right\|_{2,1},
\end{equation}
where $\hat{\mathbf{Y}}^{i,j}$ is the $j$-th predicted trajectory of pedestrian $i$, $\mathbf{Y}_i$ is the ground-truth trajectory of pedestrian $i$, and $\binom{a}{b}$ is the binomial coefficient.
The AUC metric is then computed by summing $E_K$ over all values of $K\in\llbracket 1, k\rrbracket$:
\begin{equation}
\mathrm{AUC} = \sum_{i=1}^{N} \sum_{K=1}^{k} E_K^i.
\end{equation}

All previous metrics focus on prediction accuracy. The collision rate is taken into account when assessing the social plausibility of predictions.  A collision between agents $i$ and $j$ is defined to occur at time $t$ if the predicted positions are closer than a distance threshold $\epsilon$:
\begin{equation}
\text{Collision}(i,j,t) = \mathds{1}\left[ \left\| \hat{\mathbf{y}}_t^i - \hat{\mathbf{y}}_t^j \right\|_2 < \epsilon \right],\mbox{with } i \neq j.
\end{equation}

Assuming high-quality ground truth, the threshold $\epsilon$ is set to the maximum value which produces zero collisions, considering the dataset ground truth trajectories ensuring that reported collisions correspond to genuine violations rather than natural proximity in dense crowds. 
Alternative strategies, such as using a fixed, physically interpretable threshold (e.g., $\epsilon=0.1$~m) if the scene is spatially calibrated and typical social distances calibrated for the considered crowd density, or integrating collision rates over a plausible range of thresholds, are possible, but would either risk overestimating collisions or complicate cross-dataset comparisons. The collision rate, denoted CR, is defined as:

\begin{align}
\mathrm{CR} = \frac{1}{N(N-1)\Delta T}
\sum_{t = T_{\text{obs}}+1}^{T_{\text{pred}}} \sum_{i \neq j} \text{Collision}(i,j,t).
\end{align}

In summary, ADE and AUC provide a faithful measure of accuracy over the full trajectory distribution, while FDE focuses on the goal prediction. In the results, we also report minADE and minFDE due to their popularity, as well as Collision Rate to assess social realism in multi-agent predictions.

\subsection{Results}


In Table~\ref{tab:sdd_results}, we gather multi-agent results \emph{as reported in the original papers} and use them to identify the most competitive baselines for deeper study. Prioritizing minADE as the first metric since it considers the whole trajectory, MART and TUTR appear as the SOTA.  We also note that our method is competitive: second-best minFDE ($11.78$), and among the three top for minADE. 

Now, for a more comprehensive evaluation, we consider the different metrics (cf. Section~\ref{sec:metrics}).  For fair evaluation, we retrained under the same data preprocessing the chosen models and evaluated them on SDD and on MADRAS. Table~\ref{tab:sdd_madras_combined} shows the obtained results with Y-Net, MART, TUTR and VISTA. These baselines represent distinct design paradigms: MART is a joint multi-agent model that predicts all agent futures simultaneously, while TUTR is conditional and forecasts each agent given its neighbors. Y-Net is included because VISTA is architecturally derived from it. 

\paragraph{SDD Dataset}
According to Table~\ref{tab:sdd_madras_combined}, Y-Net achieves correct minADE/minFDE ($8.05/11.99$) but demonstrates poor ADE/FDE ($32.14/60.10$) along with a high collision rate ($8.11\%$). 
These results underscore the limitations of mono-agent approaches. Among the multi-agent models, VISTA demonstrates superior overall performance, achieving lower ADE/FDE values  ($28.4/57.4$) compared to MART and TUTR, the lowest AUC ($244$), close performance to MART on minADE ($7.85$ vs. $7.60$), and the best minFDE ($11.78$). Additionally, our model's predictions are socially compliant, with zero collisions, in contrast to MART ($1.5\%$). Compared to TUTR, VISTA consistently outperforms in all error metrics.

\paragraph{MADRAS Dataset}

On the MADRAS dataset, our model achieves the best overall performance, as shown in Table~\ref{tab:sdd_madras_combined}. It yields the lowest ADE/FDE ($0.64/1.13$) as well as highly competitive minADE/minFDE ($0.18/0.25$), closely matching the best-performing model (MART) on those metrics. MADRAS dataset results confirm the benefit of multi-agent models since Y-Net shows a clear gap, with relatively high ADE/FDE ($8.74/15.29$) and a collision rate exceeding $5\%$. While its minADE/minFDE values ($0.50/0.65$) indicate that Y-Net can sometimes predict plausible trajectories, its lack of explicit interaction modeling leads to frequent socially unacceptable outcomes.
In contrast, our model achieves a very low collision rate ($0.03\%$), much lower than MART ($2.14\%$). This low value reflects our model's effective understanding of spatial constraints and interactions among agents. Compared to TUTR, our method shows significantly lower minADE/minFDE, ADE/FDE and AUC.
These results validate our model's performance across challenging scenarios.

\begin{table}[t]
  \centering
  \resizebox{\columnwidth}{!}{%
  \begin{tabular}{@{}lc | lc@{}}
    \toprule
    Method & minADE/minFDE $\downarrow$ & Method & minADE/minFDE  $\downarrow$\\
    \midrule
    PECNET~\cite{pecnet}         & 9.96/15.88 & GroupNet~\cite{groupnet}   & 9.31/16.11 \\
    MemoNet~\cite{memonet}       & 8.56/12.66 & MID~\cite{mid}             & 9.73/15.32 \\
    NSPN~\cite{nspn}             & 8.56/11.80 & DynGroupNet~\cite{dyngroupnet} & 8.42/13.58 \\
    LED~\cite{led}               & 8.48/\textbf{11.66} & MART~\cite{mart}   & \textbf{7.61}/12.19 \\
    TUTR~\cite{tutr}             & \underline{7.76}/12.69 & VISTA            & 7.85/\underline{11.78} \\
    \bottomrule
  \end{tabular}
  }
  \caption{Comparison of minADE and minFDE on the SDD dataset (in pixels). The results come directly from the articles. With this study, the top three models were selected for a more extensive comparison.}
  \label{tab:sdd_results}
\end{table}

\begin{table}[h]
  \centering
  \resizebox{\columnwidth}{!}{%
    \begin{tabular}{@{}llcccc@{}}
      \toprule
      Dataset & Method & ADE/FDE $\downarrow$ & minADE/minFDE $\downarrow$ & AUC $\downarrow$ & Collision Rate $\downarrow$ \\
      \midrule
      \multirow{4}{*}{SDD} 
             & Y-Net & 32.14/60.10 & 8.05/11.99 & 248 & 8.11\%\\
             & MART & \underline{37.6/78.5} & \textbf{7.60}/12.12 & \underline{252}  & \underline{1.50\%} \\
             & TUTR & 37.9/79.5 & \underline{7.85}/12.87 & 275 & N.A. \\
             & VISTA            & \textbf{28.4/57.4} & \underline{7.85}/\textbf{11.78} & \textbf{244} & \textbf{0\%} \\
      \midrule
      \multirow{4}{*}{MADRAS} 
             & Y-Net & 8.74/15.29 & 0.50/0.65 & 118 & 5.36\%\\
             & MART  & \underline{0.69/1.29} & \textbf{0.17/0.24} & \underline{5.65} & \underline{2.14\%} \\
             & TUTR  & 0.91/1.41 & 0.37/0.56         & 5.71 & N.A. \\
             & VISTA & \textbf{0.64/1.13} & \underline{0.18/0.25} & \textbf{5.59} & \textbf{0.03\%} \\
      \bottomrule
    \end{tabular}
  }
  \caption{Comparison on both SDD (in pixels) and MADRAS (in meters) under the same preprocessing protocol.}
  \label{tab:sdd_madras_combined}
\end{table}

\paragraph{Ablation study}

\begin{table}[h]
  \centering
  \resizebox{\columnwidth}{!}{
  \begin{tabular}{@{}cccccc c c@{}}
    \toprule
    Learnable & Fixed & Social & Temporal & Goal & $minADE/minFDE$ $\downarrow$ & Collision Rate  $\downarrow$\\
    PE & PE & Attention & Attention & Predicted & & \\
    \midrule
    \checkmark & \checkmark & \checkmark & \checkmark &  & 25.79/49.39 & N.A. \\
     &     &          & \checkmark & \checkmark & 8.55/13.09 & 4.28\% \\
     &     & \checkmark & \checkmark & \checkmark & 8.12/12.34 & 2.10\% \\
     &  \checkmark   & \checkmark & \checkmark & \checkmark & \underline{7.92/11.89} & \underline{0.75\%} \\
    \checkmark & \checkmark & \checkmark & \checkmark & \checkmark & \textbf{7.85/11.78} & \textbf{0\%} \\
    \bottomrule
  \end{tabular}
  }
  \caption{Ablation study on the effect of positional encodings (PE) and attention types on model performance. }
  \label{tab:my_label}
\end{table}

In this ablation study, we assess the contribution of the different components of our model. Table~\ref{tab:my_label} shows the impact of positional encodings (PE), social and temporal attention modules, and goal-conditioning.

Without goal-conditioning, errors increase sharply ($25.79/49.39$), confirming the goal’s role as a crucial prior for feasible long-term predictions. Then, starting with a baseline that employs only temporal attention, without any social modeling or positional encoding, this setup yields relatively high minADE/minFDE ($8.55/13.09$) and a collision rate of $4.28\%$. Introducing social attention significantly improves performance, reducing both error metrics (down to $8.12/12.34$) and collision rate (down to $2.10\%$), respectively, underscoring the value of interaction modeling. Adding fixed positional encodings further enhances temporal discrimination, lowering minADE/minFDE (down to $7.92/11.89$) and collision rate (down to $0.75\%$). Finally, combining both fixed and learnable positional encodings yields the best results, achieving $7.85/11.78$ with zero collision. This highlights the complementary benefits of sophisticated PE and full attention-based architecture. 

\paragraph{Interpreting social attention.}

The social–attention block provides, at every time step $t$, a pairwise matrix $(A^{(t)}\!\in\!\mathbb{R}^{N\times N}$ whose terms $A_{ij}^{(t)}$ quantify how strongly pedestrian $i$ (row) attends to pedestrian $j$ (column). Inspecting this matrix provides insights into which interactions the model considers critical.

\begin{figure}[ht]
    \centering
    \includegraphics[width=1.\linewidth]{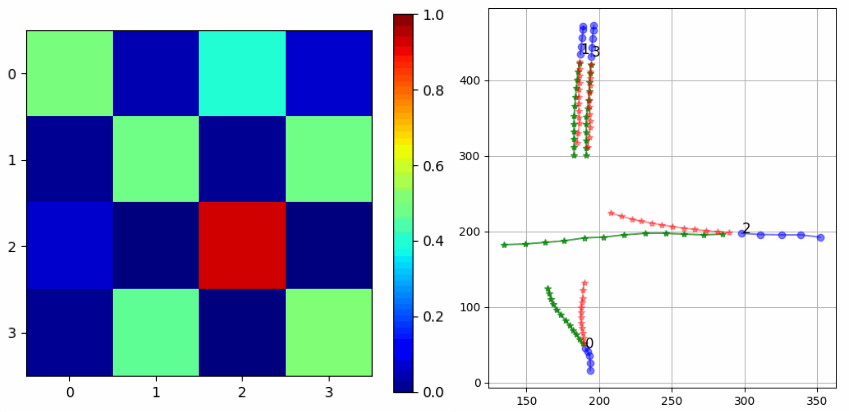}
    \caption{\textbf{SDD dataset:} the left panel shows the social-attention matrix and the right panel compares predicted (red) to future ground-truth (green) trajectories. }
    \label{fig:att}
\end{figure}

Figure~\ref{fig:att} presents a single time step prediction from the SDD dataset at $t^\star=12$. In the prediction panel on the right, pedestrians~$\mathbf{1}$ and $\mathbf{3}$ move in parallel and close proximity, indicating group formation at this time point. The attention map on the left at $t^\star$ supports this observation: the cross-weight values $A^{(t^\star)}_{1,3}$ and $A^{(t^\star)}_{3,1}$ are high, as the self-weights $A^{(t^\star)}_{1,1}$ and $A^{(t^\star)}_{3,3}$, reflecting focused within-group attention. Regarding pedestrians $\mathbf{0}$ and $\mathbf{2}$, pedestrian $\mathbf{0}$ allocates significant attention to $\mathbf{2}$, while the reverse is less pronounced. Pedestrian $\mathbf{2}$ demonstrates dominant self-attention, $A^{(t^\star)}_{2,2}$. This asymmetry corresponds to the spatial configuration shown in the prediction panel: pedestrian $\mathbf{0}$ pays attention to $\mathbf{2}$, because of its speed, whereas $\mathbf{2}$ proceeds independently and focuses primarily on itself.

Supplementary Material's Figure~1 illustrates a prediction from the MADRAS dataset. Initially, pedestrians $\mathbf{0}{-}\mathbf{6}$ assign noticeable weights to pedestrians $\mathbf{7}$ and $\mathbf{8}$ because of their spatial proximity. However, after $4-5$ frames, those weights drop sharply: $A_{0{:}6,\,7}^{(t)}$ and $A_{0{:}6,\,8}^{(t)}\downarrow$, indicating that the model has inferred that agents $\mathbf{7}$ and $\mathbf{8}$ are diverging from the main group. The predicted trajectories in the right panel support this interpretation: pedestrians $\mathbf{7}$ and $\mathbf{8}$ indeed veer to the right, leaving the main group.  

At the beginning of the sequence ($t\!=\!T_{obs}+1$), the second row of the attention matrix is nearly uniform, meaning that pedestrian~$\mathbf{1}$ distributes attention evenly across nearby agents due to proximity. Over time, this distribution changes: the weights gradually migrate toward the diagonal $A_{1,1}$ (strong self–attention) and toward column~10, i.e.\ $A_{1,10}$.  The model has identified pedestrian~$\mathbf{10}$ as the only agent on a plausible collision with pedestrian~$\mathbf{1}$, hence it adjusts its attention accordingly.

Visualizing the pairwise attention matrices alongside the predicted trajectories makes VISTA interpretable. The gradual reallocation of weights reflects the dynamics of the scene and provides a clear rationale for each prediction. 

\paragraph{Long-term prediction}
To go further, we evaluated our model for long-term prediction. We report its performance on the 5+30 prediction setting of SDD in Table~\ref{tab:sdd_comp_long}. 

Here, the considered metrics are minADE/minFDE, Collision Rate, as previously, but also KDE-NLL, MissRate to align with~\cite{das2024distilling}.  Kernel Density Estimation Negative Log-Likelihood~\cite{trajectron} evaluates how well the predicted distribution aligns with the ground truth trajectory. MissRate~\cite{towards} quantifies the percentage of forecasts that deviate from the ground truth by more than a fixed threshold.

Compared to SOTA methods, VISTA demonstrates competitive results, achieving the best scores in terms of minFDE, KDE-NLL,  and MissRate, while slightly lagging on minADE with respect to Di-Long. For Collision Rate, VISTA also achieves a substantially lower value ($0.76\%$) than Y-Net ($1.45\%$), confirming that our recursive goal–trajectory fusion and social attention design yield forecasts that are not only accurate but also safer in dense multi-agent settings. These results underline the potential of extending our approach to longer horizons and suggest that specific adaptations to capture long-term dependencies could further improve its performance in this challenging setting.

\begin{table}[t]
  \centering
  \resizebox{\columnwidth}{!}{%
  \begin{tabular}{@{}lc ccc@{}}
    \toprule
        Method & minADE/minFDE $\downarrow$ & KDE-NLL $\downarrow$ & MissRate $\downarrow$ & Collision Rate $\downarrow$\\
        \midrule
        Y-Net~\cite{YNet2021} & \textbf{49.61}/66.41 & 9.10 & 0.14 & 1.45\%\\
        Di-Long~\cite{das2024distilling} & 48.21/72.41 & 10.85 & 0.28 & N.A. \\
        VISTA & 50.72/\textbf{65.13} & \textbf{9.09} & \textbf{0.002} & \textbf{0.76}\%\\
        \bottomrule
  \end{tabular}
  }
  \caption{Long-term trajectory prediction results on the SDD dataset. $\ast$ means that we retrained the model to align the data pre-processing.}
  \label{tab:sdd_comp_long}
\end{table}

\section{Conclusion}
\label{sec:conclusion}

In conclusion, this paper introduced VISTA, an innovative architecture for multi-agent trajectory prediction, based on recursive goal-conditioned transformers. By integrating attentive fusion of goals with past trajectories, fine-grained social interaction modeling through token-based attention, and interpretable pairwise attention maps that reveal agent interactions and mutual influences, VISTA significantly enhances current performance benchmarks. 

The results obtained from the MADRAS and SDD datasets demonstrate VISTA's remarkable effectiveness, notably reducing collision rates to nearly zero levels while outperforming existing models in terms of ADE/FDE metrics.

Moreover, the built-in pairwise attention mechanisms provide unprecedented transparency into model decisions, enhancing its suitability for deployment in safety-critical contexts. These advancements pave the way for robust, socially aware applications within autonomous systems.

\newpage
{
    \small
    \bibliographystyle{ieeenat_fullname}
    \bibliography{main}
}

\end{document}